%% file: main.tex
\documentclass[11pt]{article}
\usepackage{naacl2021}
\usepackage{microtype}
\usepackage{times}
\usepackage{latexsym}
\usepackage[T1]{fontenc}
\usepackage{color}
\usepackage{xcolor}

\usepackage{amsmath}
\usepackage{amsfonts}       %
\usepackage{amssymb}
\usepackage{caption}
\usepackage{subcaption}

\usepackage{xspace}
\usepackage{latexsym}

\usepackage{textcomp}
\usepackage{graphicx}
\usepackage{hyperref}
\usepackage{float}
\restylefloat{table}

\usepackage{tabularx}
\usepackage{multirow}
\usepackage{booktabs}       %
\usepackage{arydshln}
\usepackage{nicefrac}       %
\usepackage{microtype}      %

\usepackage{verbatim}
\usepackage[utf8]{inputenc}

\usepackage{tabu}

\DeclareMathOperator*{\argmax}{arg\,max}

\newcommand{\rudy}[1]{}
\newcommand{\daniel}[1]{}
\newcommand{\coline}[1]{}
\newcommand{\dan}[1]{}
\newcommand{\trevor}[1]{}

\newcommand{\subtask}[0]{subgoal\xspace}
\newcommand{\subtasks}[0]{subgoals\xspace}

\newcommand{\module}[0]{module\xspace}
\newcommand{\modules}[0]{modules\xspace}

\newcommand{\Module}[0]{Module\xspace}
\newcommand{\Modules}[0]{Modules\xspace}

\newcommand\eg{\emph{e.g.}\ }

\newcommand\goto{\textsc{GoTo}\xspace}

\title{Modular Networks for Compositional Instruction Following}

\author{Rodolfo Corona ~ Daniel Fried ~ Coline Devin ~ Dan Klein ~ Trevor Darrell\\
UC Berkeley \\
\texttt{\{rcorona,dfried,coline,klein,trevordarrell\}@berkeley.edu}%
}

\begin{document}
\maketitle
\begin{abstract}

Standard architectures used in instruction following often struggle on novel compositions of \emph{subgoals} (e.g.\ navigating to landmarks or picking up objects) observed during training.
We propose a modular architecture for following natural language instructions that describe sequences of diverse subgoals. In our approach, subgoal modules each carry out natural language instructions for a specific subgoal type. A sequence of modules to execute is chosen by learning to segment the instructions and predicting a subgoal type for each segment. When compared to standard, non-modular sequence-to-sequence approaches on ALFRED \cite{shridhar:cvpr20}, a challenging instruction following benchmark, we find that modularization improves generalization to novel subgoal compositions, as well as to environments unseen in training. 

\end{abstract}

\input{figure1.tex}
\input{intro_and_related.tex}

\input{methods.tex}

\input{experiments.tex}
\input{results.tex}
\input{conclusion.tex}

\section*{Acknowledgments}
This material is based upon work supported by the National Science Foundation Graduate Research Fellowship Program under Grant No. DGE 1752814, a Ford Foundation fellowship to the first author, a Google PhD fellowship to the second author, and by DARPA through the XAI program and the LwLL program.

\bibliographystyle{acl_natbib}
\bibliography{main}

\clearpage

\input{appendix.tex}

\end{document}

%% file: figure1.tex
\begin{figure}[t]
    \centering
    \includegraphics[width=\linewidth]{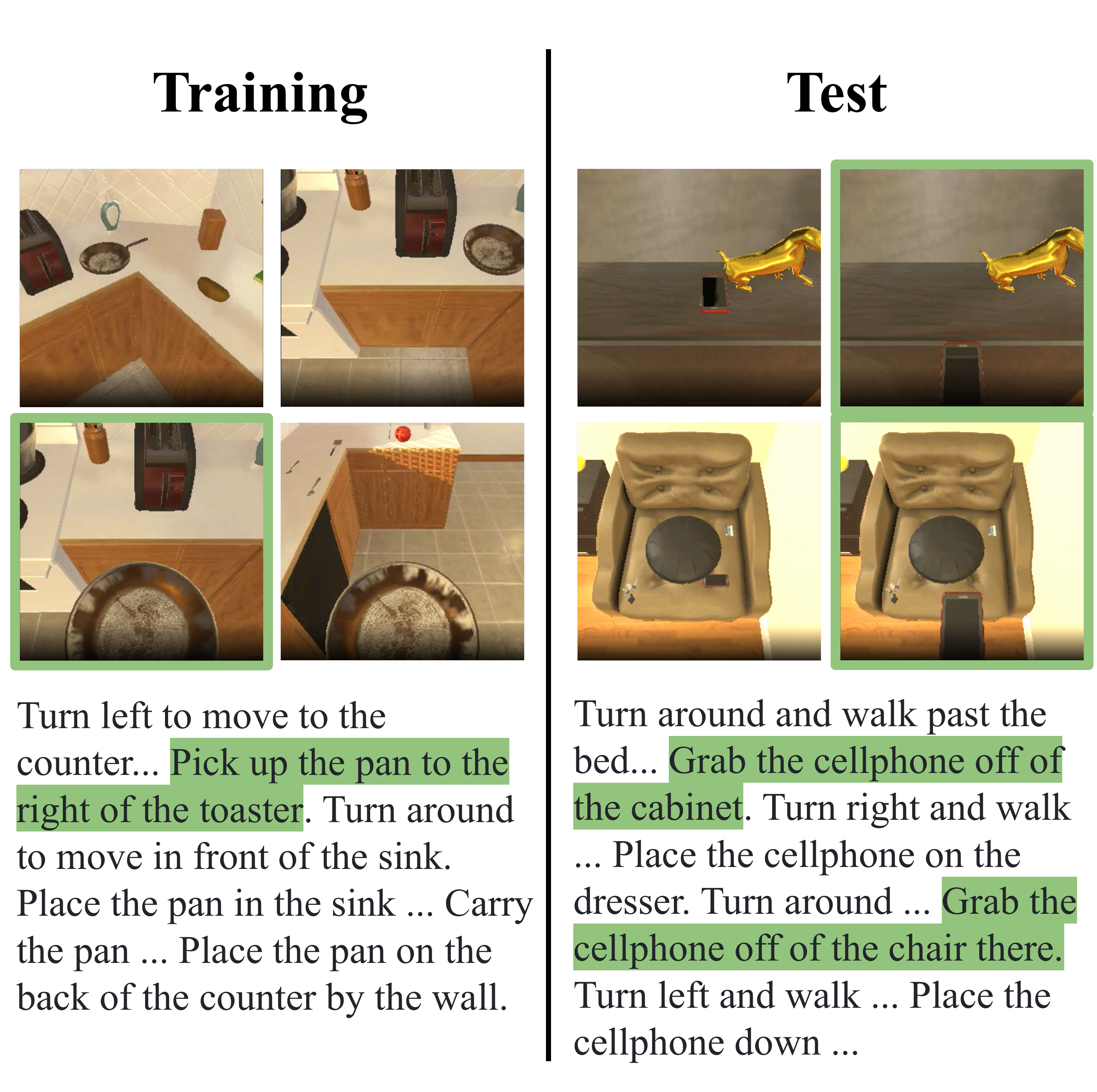}
    \caption{At evaluation time, an instruction following agent may need to generalize both to novel chains of subgoals encountered during training as well as to completely new environments. In the generalization condition above, the agent must generalize to \textit{multiple} pickup actions (in green) at test time, whereas only single ones were seen at training, as well as to a new house. We propose a modular architecture to handle these cases.}
    \label{fig:generalization_conditions}
\end{figure}

%% file: intro_and_related.tex
\section{Introduction}

Work on grounded instruction following \cite{macmahon2006walk,vogel2010learning,tellex2011understanding,chen2011navigation,artzi2013instructions}
has recently been driven by sequence-to-sequence models \cite{Mei16Instructions,hermann2017grounded}, which allow end-to-end grounding of linguistically-rich instructions into equally-rich visual contexts \cite{misra2018mapping,anderson2018vision,chen2019touchdown}.
These sequence-to-sequence models are \emph{monolithic}:
they consist of a single network structure which is applied identically to every example in the dataset. 

Monolithic instruction following models typically perform well when evaluated on test data from the same distribution seen during training.
However, they often struggle in 
\textit{compositional generalization}: composing atomic parts, such as actions or goals, where the parts are seen in training but their compositions are not \cite{lake2017generalization,ruis2020benchmark,hill2020environmental}.

In this work, we improve compositional generalization in instruction following with 
\emph{modular networks}, which have been successful in non-embodied language grounding tasks \citep{andreas2016neural,hu2017learning,cirik2018using,yu2018mattnet,mao2019neuro,han2019visual} and in following synthetic instructions or symbolic policy descriptions \citep{andreas2017modular,oh_2017,das2018neural}.
Modular networks split the decision making process into a set of neural modules. Modules are each specialized for some function, composed into a structure specific to each example, and trained jointly to complete the task.

\input{figure2}

Prior work has found that modular networks often perform well in compositional generalization because of their composable structure \citep{devin2017learning,andreas2017modular,bahdanau2018systematic,purushwalkam2019task}, and that they can generalize to new environments or domains through module specialization \cite{hu_19,blukis20a}.
However, all this work has either focused on grounding tasks without a temporal component or used a network structure which is not predicted from language.

We propose a modular architecture for embodied vision-and-language instruction following\footnote{Code and dataset splits may be found at \href{https://github.com/rcorona/modular_compositional_alfred}{github.com/rcorona/modular\_compositional\_alfred}.}, and find that this architecture improves generalization on unseen compositions of \emph{\subtasks} (such as navigation, picking up objects, cleaning them, etc.). %
We define separate sequence-to-sequence modules per type of subgoal. These modules are strung together to execute complex high-level tasks. 
We train a controller to predict a sequence of subgoal types  from language instructions, which determines the order in which to execute the modules.

We evaluate models on the ALFRED dataset~\citep{shridhar:cvpr20}, an instruction-following benchmark containing a diverse set of household tasks. We focus on compositional generalization: carrying out instructions describing novel high-level tasks, containing novel compositions of actions (see Figure~\ref{fig:generalization_conditions} for an example).
We find that our modular model improves performance on average across \subtask types when compared to a standard, monolithic sequence-to-sequence architecture.
Additionally, we find improved generalization to environments not seen in training.

%% file: figure2.tex
\begin{figure*}[t!]
\includegraphics[width=\textwidth]{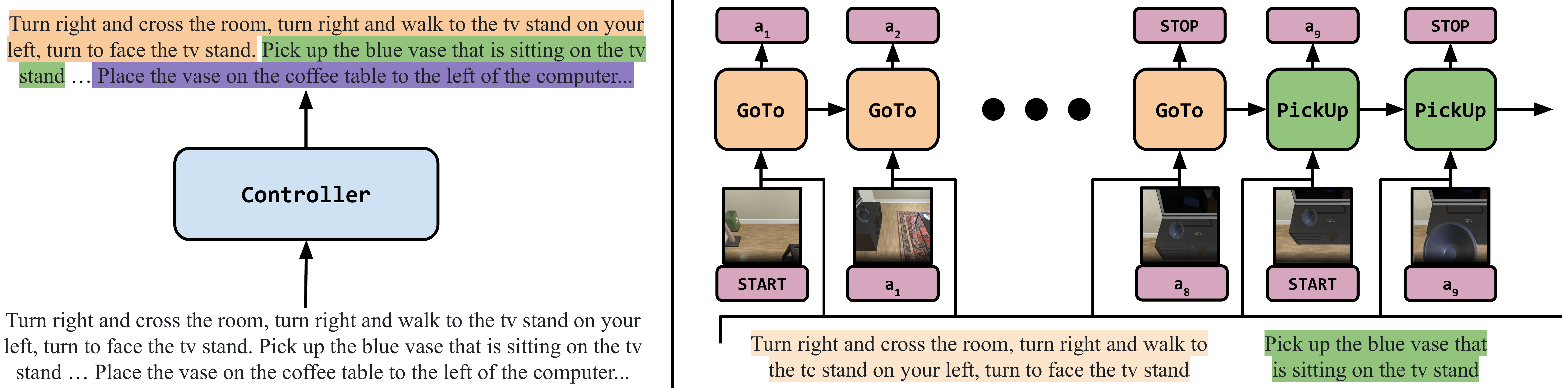}
\caption{
\label{fig:teaser}
Our modular approach first uses a \emph{controller} (left) trained with supervised learning to segment a given instruction and label segments with subgoal types (\eg \goto, \textsc{Pickup}) to execute. These subgoal types are used to chain together \emph{\modules} (right) to carry out instructions in the environment. Each \module  is a separately-parameterized sequence-to-sequence model that conditions on an attended representation of the instruction sequence, the visual observations, and the action taken at the previous timestep. \Modules pass recurrent hidden states to each other.
}
\end{figure*}

%% file: methods.tex
\section{Modular Instruction Following Networks}
\label{sec:method}
We focus on following instructions in embodied tasks involving navigation and complex object interactions, as shown in Figure~\ref{fig:teaser}.

In training, each set of \emph{full instructions} (\eg ``Turn right and cross the room ... Place the vase on the coffee table to the left of the computer.'')
is paired with
a \emph{demonstration} of image observations and
actions.
\daniel{I took out notation since we no longer use it, and object interaction masks (which are mentioned later, and not explicitly in the figure}
In training, we further assume that the full instruction is segmented into \emph{\subtask instructions}, and each \subtask instruction is labeled with one of a small number (in our work, 8) of \emph{\subtask types} , \eg [``Walk to the coffee maker.'': \goto], [``Pick up the dirty mug...'': \textsc{PickUp}], \dots, and paired with the corresponding segment of the demonstration.

During evaluation, the agent is given only full instructions (which are unsegmented and unlabeled),
and must predict a sequence of actions to carry out the instructions, conditioning on the image observations it receives. 

Our modular architecture for compositional instruction following consists of a \emph{high-level controller} (Figure~\ref{fig:teaser}, left), and \emph{modules} for each \subtask type (Figure~\ref{fig:teaser}, right).
The high-level controller chooses \modules to execute in sequence based on the natural language instructions, and each chosen \module executes until it outputs a STOP action. The \modules all share the same sequence-to-sequence architecture, which is the same as the monolithic architecture. We initialize each module's parameters with parameters from the monolithic model, and then fine-tune the parameters of each module to specialize for its \subtask.

\subsection{Instruction-Based Controller}
\label{sec:controller}

Our instruction-based controller is trained to segment a full instruction into sub-instructions and predict the \subtask type for each sub-instruction. %
We use a linear chain CRF \citep{lafferty2001conditional} that conditions on a bidirectional-LSTM encoding of the full instruction and predicts tags for each word, which determine the segmentation and sequence of \subtask types. 
This model is based on standard neural segmentation and labelling models \citep{huang2015bidirectional,lample2016ner}.

We train the controller on the ground-truth instruction segmentations and \subtask sequence labels, and in evaluation use the model to predict segmentations and their associated \subtask sequences (Figure~\ref{fig:teaser}, top left). 
This predicted sequence of \subtasks determines the order to execute the \modules (Figure~\ref{fig:teaser}, right).
We use a BIO chunking scheme to jointly segment the instruction and predict a subgoal label for each segment.

Formally, for a full instruction of length $N$, the controller defines a distribution over subgoal tags $s_{1:N}$ for each word given the instruction $x_{1:N}$ as
\begin{equation*}
\label{eq:subtask_labels}
p(s_{1:N} \mid x_{1:N}) \propto \exp \sum_{n=1}^N \left (U_{s_n} + B_{s_{n-1},s_{n}} \right)
\end{equation*}
The \subtask tag scores $U_{s_n}$ for word $n$ are given by a linear projection of bidirectional LSTM features for the word at position $n$. %
The tag transition scores $B_{s_{n-1},s_{n}}$ are learned scalar parameters.

In training, we supervise $s_{1:N}$ using the segmentation of the instruction $x_{1:N}$ into $K$ \subtask instructions and the \subtask label for each instruction. 
To predict \subtasks for a full instruction in evaluation, we obtain $\argmax_{s_{1:N}} p(s_{1:N} \mid x_{1:N})$ using Viterbi decoding, which provides a segmentation into sub-instructions and a \subtask label for each sub-instruction.

The controller obtains 96\% exact match accuracy on \subtask sequences on validation data. 

\subsection{\Module Architecture}\label{sec:subgoal_modular}

Our modularized architecture may be seen in Figure \ref{fig:teaser}, right.
The architecture consists of 8 independent modules, one for each of the 8 subgoals in the domain (\eg \textsc{GoTo}, \textsc{PickUp}). %
For each \module, we use the same architecture as \citet{shridhar:cvpr20}'s monolithic model. This is a sequence-to-sequence model composed of an LSTM decoder taking as input an attended embedding of the natural language instruction, pretrained ResNet-18~\citep{he2016deep} features of the image observations, and the previous action's embedding.
Hidden states are passed between the \modules' LSTM decoders at subgoal transitions (Figure~\ref{fig:teaser}, right). 

At each time step, each module 
$M^i$
computes its hidden state based on the last time step's action $a_{t-1}$, the current time step's observed image features $o_t$, an attended language embedding $\hat{x}_t^i$, and the previous hidden state $h_{t-1}^i$: 
\begin{align*}
e_t^i &= [a_{t-1} ; o_t ; \hat{x}_t^i] \\
h_t^i &= \text{LSTM}_i(e_t^i, h_{t-1}^i) 
\end{align*}
Each module's attended language embedding $\hat x_t^i$ is produced using its own attention mechanism over embeddings $X = x_{1:N}$ of the language instruction, which are produced by a bidirectional LSTM encoder:
\begin{align*}
    z_t^i &= (W_x^i h_{t-1}^i)^\top X \\
    \alpha _t^i &= \text{Softmax}(z_t^i) \\
    \hat{x}_t^i &= (\alpha_t^i)^\top X
\end{align*}
Finally, the action $a_t$ and object interaction mask $m_t$ are predicted from $h_t^i$ and $e_t^i$ with a linear layer and a deconvolution network respectively. 
More details about this architecture can be found in \citet{shridhar:cvpr20}.
Both the action and mask decoders, well as the language encoder, are shared across modules.%
\footnote{The modules' instruction encoder is separate from the controller's encoder (Sec.~\ref{sec:controller}), as we found it possible to achieve high performance on the \subtask prediction task using a smaller encoder than the one used by the modules.}
\daniel{details about which parameters are shared, in appendix?}

Our use of subgoal modules is similar to the hierarchical policy approaches of \citet{andreas2017modular}, \citet{oh_2017}, and \citet{das2018neural}.
However, in those approaches, the input to each module is symbolic (\eg \textsc{find[kitchen]}). In contrast, all modules in our work condition directly on natural language.

\subsection{Training}

We first pre-train the monolithic model by maximizing the likelihood of the ground-truth trajectories in the training data \cite{shridhar:cvpr20}. 
We train for up to 20 epochs using the Adam optimizer \citep{kingma2014adam} with early stopping on validation data (see Appendix~\ref{app:hyperparameters} for hyperparameters).
We use this monolithic model to initialize the parameters of each of the \modules, which have identical architecture to the monolithic model, and fine-tune them using the same training and early stopping procedure on the same validation data,\footnote{Additionally, we append a special STOP action to the end of each \module's action sequence so that it can predict when to give control back to the high-level controller.}
allowing the monolithic model's parameters to specialize for each module. 
Each module predicts only the actions for its segment of each trajectory; however, modules are jointly fine-tuned, passing hidden states (and gradients) from module to module.

%% file: experiments.tex
\section{Generalization Evaluation}
We evaluate models on out-of-domain generalization in two conditions (see below) using the ALFRED benchmark~\cite{shridhar:cvpr20}, comparing our modular approach to their non-modular sequence-to-sequence model.
ALFRED is implemented in  AI2-THOR 2.0 \cite{kolve2017ai2}, which contains a set of simulated environments with realistic indoor scene renderings and object interactions.

The dataset contains approximately 25K expert instruction-trajectory pairs, comprised of 3 instructions for each of 8K unique trajectories. The instructions include both a high level instruction and a sequence of low level instructions. 
In our experiments, we do not use the high level instructions, which \citet{shridhar:cvpr20} found to produce comparable results when evaluated on generalization to unseen environments with these architectures.

Figure~\ref{fig:generalization_conditions} shows two example trajectories and their associated instructions. %
Trajectories are composed (see Sec.~\ref{sec:method}) of sequences of eight different types of \subtasks: navigation (\goto) and a variety of object interactions (\eg \textsc{PickUp}, \textsc{Clean}, \textsc{Heat}). 
Each \subtask's subtrajectory is composed of a sequence of low-level discrete actions which specify commands for navigation or object interactions (which are accompanied by image segmentations to choose the object to interact with).

\subsection{Generalization Conditions}

The ALFRED dataset was constructed to test generalization to novel instructions and unseen environments.
However, all evaluation trajectories in the dataset correspond to sequences of \subtasks that are seen during training. For example, some training and evaluation instances might both correspond to the underlying \subtask sequence \textsc{GoTo}, \textsc{PickUp}, \textsc{GoTo}, \textsc{Put}, but differ in their low-level actions, their language descriptions, and possibly also the environments they are carried out in. 

\paragraph{Novel Tasks.}
We evaluate models' ability to generalize to different high-level tasks (compositions of \subtasks) than seen in training. 
\rudy{Is the distinction between "subgoals" and "tasks" clear to the reader?} \daniel{I tried to address} 
The dataset contains seven different task types, such as \emph{Pick \& Place}, as described in Appendix~\ref{app:tasktypes}. %
We hold out two task types and evaluate models on their ability to generalize to them: \emph{Pick Two \& Place} and \emph{Stack \& Place}.
These tasks are chosen because they contain \subtask types that are all individually seen in training, but typically in different sequences.

We create generalization splits \emph{pick-2-seen} and \emph{pick-2-unseen} by filtering the \emph{seen} and \emph{unseen} splits below to contain only \emph{Pick Two \& Place} tasks, and remove all \emph{Pick Two \& Place} tasks from the training data. We create splits \emph{stack-seen} and \emph{stack-unseen} for \emph{Stack \& Place} similarly.

\paragraph{Novel Instructions and Environments}

This is the standard condition defined in the original ALFRED dataset. There are two held-out validation sets: \emph{seen}, which tests generalization to novel instructions and trajectories but through environments seen during training, and \emph{unseen}, which tests generalization to novel environments: rooms with new layouts, object appearances, and furnishings. %

%% file: results.tex
\subsection{Results}
\input{table.tex}

We compare our modular architecture with the monolithic baseline, averaging performance over models trained from 3 random seeds. For each generalization condition, we measure success rates over full trajectories as well as over each subgoal type independently. Due to the challenging nature of the domain, subgoal evaluation provides finer-grained comparisons than full trajectories. %

We use the same evaluation methods and metrics as in \citet{shridhar:cvpr20}.
Success rates are weighted by path lengths to penalize successful trajectories which are longer than the ground-truth demonstration trajectory. To evaluate full trajectories, we measure path completion: the portion of subgoals completed within the full trajectories.
To evaluate the subgoals independently, we advance the model along the expert trajectory up until the point where a given subgoal begins (to maintain a history of actions and observations), then require the model to carry out the subgoal from that point. %

We also report results from \citet{shridhar:cvpr20} and \citet{singh2020moca}. We note that the approach of \citet{singh2020moca} obtains higher performance on full trajectories than the system of \citet{shridhar:cvpr20} (which we base our approach on) primarily by introducing a modular object interaction architecture (shared across all \subtasks) and a pre-trained object segmentation model. These techniques could also be incorporated into our approach, which uses modular components for individual \subtask types.

\paragraph{Novel Tasks.}
Table~\ref{tab:subtask_abblations} shows for each split the success rates on \subtasks appearing in at least 50 validation examples.
The modular outperforms the monolithic model on both seen and unseen splits (Tables~\ref{tbl:pick_2_split} and \ref{tbl:stack_split}). %
Full trajectory results for novel task generalization are shown in Table~\ref{tab:standard}. 
In the double generalization condition (unseen environments for the held-out pick-2 and stack tasks) on full trajectories, neither model completes \subtasks successfully.
Overall, we find that modularity helps across most generalization conditions. 

\paragraph{Generalization to novel environments.}\label{sec:independent_subgoal_experiments}
We also compare models on generalization to unseen environments.
In the independent \subtask evaluation,
the monolithic and modular models perform equally on average in the standard-seen split (Table~\ref{tbl:standard_split}, top). However, in the standard-unseen split (Table~\ref{tbl:standard_split}, bottom), our modular model outperforms the baseline substantially, with an average success rate of $57\%$ compared to the monolithic model's $46\%$. (On \subtask types not shown, the modular model still outperforms the monolithic, by margins up to 16\%.) In the full trajectory results (Table~\ref{tab:standard}) we see comparable performance between the monolithic and modular models on unseen environments.

\begin{table}
\centering
\small

\begin{tabular}{rcccc}
& Standard & Standard & Pick-2 & Stack \\
Model & seen & unseen & seen & seen \\
\toprule
S+ & 9.4 (5.7) & 7.4 (4.7) & --- & --- \\
MOCA & 28.5 (22.3) & 13.4 (8.3) & --- & --- \\
\hdashline \\[-0.8em]
Mono. & \textbf{10.9 (7.0)} & \textbf{7.1} (4.9) & 1.3 (1.6) & 1.3 (0.3) \\
Mod. & 9.1 (6.6) & 7.0 \textbf{(5.5)} & \textbf{1.5} (1.6) & \textbf{1.5 (0.4)} \\
\bottomrule
\end{tabular}
\caption{\label{tab:standard}We compare performance of the monolithic and modular models on full trajectories, reporting the percentages of \subtasks correctly completed. Numbers in parentheses weight these percentages by path length. S+ gives results from \citet{shridhar:cvpr20}, and MOCA from \citet{singh2020moca}. %
\vspace{-1em}
}
\end{table}

%% file: table.tex
\begin{table}[t]
\setlength{\belowrulesep}{0.8ex}
\begin{center}
\begin{subtable}[t]{\linewidth}
\resizebox{\columnwidth}{!}{
\begin{small}
\begin{tabular}{r@{\hspace{4pt}}r@{\hspace{10pt}}l@{\hspace{7pt}}l@{\hspace{7pt}}l@{\hspace{7pt}}l@{\hspace{7pt}}l@{\hspace{7pt}}l@{\hspace{7pt}}l@{\hspace{7pt}}l@{\hspace{7pt}}|l@{\hspace{7pt}}}
&Model & \rotatebox{75}{Clean} & \rotatebox{75}{Cool} & \rotatebox{75}{Goto}  & \rotatebox{75}{Heat} & \rotatebox{75}{Pickup} & \rotatebox{75}{Put} & \rotatebox{75}{Slice} & \rotatebox{75}{Toggle} & \rotatebox{75}{Avg.} \\ %
\toprule
\multirow{4}{*}{\rotatebox{90}{{seen~}}}
& S+ & 82 & 87 & 49 & 85 & 32 & 80 & 23 & 97 & 67 \\
& MOCA & 79 & 87 & 54 & 84 & 53 & 62 & 51 & 93 & 70 \\
\hdashline & & & & & & & & & \\[-0.8em]
& Mono. & 82 & \textbf{88$^*$} & \textbf{52$^{**}$} & \textbf{82$^{**}$} & 37 & \textbf{81} & 34 & \textbf{98} & \textbf{69$^{*}$} \\ %
& Mod. & 82 & 86 & 43 & 80 & \textbf{41$^{**}$} & 80 & \textbf{37} & 97 & 68 \\ %
\midrule
\multirow{4}{*}{\rotatebox{90}{{unseen}}}
& S+ & 21 & 94 & 21 & 88 & 20 & 51 & 14 & 54 & 45 \\
& MOCA & 71 & 38 & 32 & 86 & 44 & 39 & 55 & 11 & 47\\
\hdashline & & & & & & & & & \\[-0.8em]
& Mono. & 28 & 90 & \textbf{23$^{**}$} & \textbf{89$^{**}$} & 25 & 48 & 25 & 42 & 46 \\ %
& Mod. & \textbf{67$^{**}$} & \textbf{94$^*$} & 14 & 85 & \textbf{28$^{**}$} & \textbf{55} & \textbf{39$^*$} & \textbf{73$^*$} & \textbf{57$^{**}$} \\ %
\bottomrule
\end{tabular}
\end{small}
}
\caption{\label{tbl:standard_split}Standard (novel environments) validation splits}
\end{subtable}

\begin{subtable}[t]{0.49\linewidth}
\resizebox{\columnwidth}{!}{
\begin{small}
\begin{tabular}{r@{\hspace{4pt}}r@{\hspace{10pt}}l@{\hspace{7pt}}l@{\hspace{7pt}}l@{\hspace{7pt}}|l@{\hspace{7pt}}}
&Model & \rotatebox{75}{Goto} & \rotatebox{75}{Pickup} & \rotatebox{75}{Put} & \rotatebox{75}{Avg.} \\ %
\toprule
\multirow{2}{*}{\rotatebox{90}{{seen~}}}
& Mono. & \textbf{18$^*$} & 21 & 48 & 29 \\[0.6ex] %
& Mod. & 14 & \textbf{35$^*$} & \textbf{63$^*$} & \textbf{37$^*$} \\[0.5ex] %
\midrule
\multirow{2}{*}{\rotatebox{90}{{unseen}}}
& Mono. & 14 & 12 & 14 & 13 \\[0.6ex] %
& Mod. & 14 & \textbf{25$^{**}$} & \textbf{33$^{**}$} & \textbf{24$^{**}$} \\[0.5ex] %
\bottomrule
\end{tabular}
\end{small}
}
\caption{\label{tbl:pick_2_split}Pick-2 task splits}
\end{subtable}
\begin{subtable}[t]{0.49\linewidth}
\resizebox{\columnwidth}{!}{
\begin{small}
\begin{tabular}{l@{\hspace{4pt}}r@{\hspace{10pt}}l@{\hspace{7pt}}l@{\hspace{7pt}}l@{\hspace{7pt}}|l@{\hspace{7pt}}}
&Model & \rotatebox{75}{Goto} & \rotatebox{75}{Pickup} & \rotatebox{75}{Put} & \rotatebox{75}{Avg.} \\ %
\toprule 
\multirow{2}{*}{\rotatebox{90}{{seen~}}}
& Mono. & \textbf{28$^*$} & 15 & 54 & 32 \\[0.6ex] %
& Mod. & 21 & \textbf{27$^*$} & \textbf{58$^*$} & \textbf{35$^*$} \\[0.5ex] %
\midrule
\multirow{2}{*}{\rotatebox{90}{{unseen}}}
& Mono. & \textbf{19} & \phantom{0}7 & 25 & 17 \\[0.6ex] %
& Mod. & 14 & \textbf{16$^*$} & \textbf{28} & \textbf{19$^*$} \\[0.5ex] %
\bottomrule
\end{tabular}
\end{small}
}
\caption{\label{tbl:stack_split}Stack task splits}
\end{subtable}
\end{center}
\vspace{-0.5em}
\caption{\label{tab:subtask_abblations}Path weighted \subtask success percentages, by \subtask type, on the various generalization splits, and averaged across \subtask types (\emph{Avg.}). %
We compare the performance of the monolithic (\emph{Mono.}) model to our modular model (\emph{Mod.}). %
The modular model generalizes better on average to
unseen environments (standard-unseen) and to both seen and unseen environments for two held-out task types: \emph{Pick-2} and \emph{Stack}. 
Bolded numbers show the best model between Mono and Modular, with $^*$ and $^{**}$ denoting differences that are statistically significant at the $p < 0.15$ and $p < 0.05$ levels, respectively, by a one-tailed t-test.  
S+ gives results from \citet{shridhar:cvpr20} and MOCA from \citet{singh2020moca}.
\vspace{-1em}
}
\end{table}

%% file: conclusion.tex
\section{Conclusions}
We introduced a novel modular architecture for grounded instruction following where each module is a sequence-to-sequence model conditioned on natural language instructions. With the ALFRED dataset as a testbed, we showed that our modular model achieves better out-of-domain generalization, generalizing better at the \subtask level to 
novel task compositions and unseen environments than the monolithic model used in prior work. 
All of the module types in our model currently use separate parameterizations but identical architectures; future work might leverage the modularity of our approach by using specialized architectures, training procedures, or loss functions for each subgoal type.
Furthermore, unsupervised methods for jointly segmenting instructions and trajectories without requiring labeled subgoal labels and alignments would be a valuable addition to our framework.

%% file: appendix.tex
\appendix
\section{Implementation Details}
\label{app:implementation}

\subsection{Model and Training Hyperparameters}
\label{app:hyperparameters}
We list the hyperparameters used for all models in Table \ref{tab:hyperparams}, we refer the reader to \citep{shridhar:cvpr20} for more details on the usage of each hyperparameter. 
Submodules are each structured identically to the monolithic baseline (e.g. each one had a 512 dimensional hidden state). 

\begin{table}[]
    \centering
    \begin{tabular}{cc}
        \textbf{Hyperparameter} & \textbf{Value} \\
        \toprule
        Optimizer & Adam \\
        Learning Rate & 1e-4\\
        Batch Size & 8 \\
        Hidden State Dim & 512 \\
        Word/Action Embedding Dim & 100 \\
        Zero-Goal & True \\
        Zero-Instr & False \\
        Lang. Dropout & 0.0 \\
        Vision Dropout & 0.0 \\
        Input Dropout & 0.0 \\ 
        Attn. Dropout & 0.0 \\
        Actor Dropout & 0.0 \\
        LSTM Dropout & 0.3 \\
        Mask Loss Wt. & 1.0 \\
        Action Loss Wt. & 1.0 \\
        \bottomrule
    \end{tabular}
    \caption{Model hyperparameters. These settings largely follow the default parameters used by \citet{shridhar:cvpr20}.}
    \label{tab:hyperparams}
\end{table}

\subsection{Hardware and Training Times}

Models were trained on a Quadro RTX 6000 24GB GPU running on a machine with a 14 core Intel Xeon Gold 5120 CPU, with a runtime of approximately 14 hours. 
Evaluation was done on a V100 16GB GPU on a machine with a 4-core CPU. 
Subgoal evaluation took approximately 8 hours per split, and full trajectory evaluation approximately 1 hour. 

\subsection{Evaluation}

We evaluate our model using the evaluation code provided by \citet{shridhar:cvpr20}.\footnote{https://github.com/askforalfred/alfred}

\section{ALFRED Dataset Details}
\label{app:dataset}
\daniel{fill in some more details here, following the reproducibility criteria: https://2020.emnlp.org/call-for-papers}

In ALFRED, the agent observes a first person view, navigates with discrete grid movement, and uses objects by outputting a segmentation mask over its image observation. 
The dataset contains approximately 25K expert instruction-trajectory pairs, pertaining to about 8K unique trajectories.

\subsection{Task Types}
\label{app:tasktypes}
The dataset contains demonstrations for 7 different kinds of tasks.
\paragraph{Pick \& Place} The agent must pickup a specified object, bring it to a destination, and place it. For example, ``Pick up a vase, place it on the coffee table."

\paragraph{Examine in Light} The agent must pickup an object and bring it to a light source. For example, ``Examine the remote control under the light of the floor lamp
."

\paragraph{Heat \& Place} The agent must pickup an object, put it in the microwave, toggle the microwave, take the object out of the microwave, and finally place the heated object at a specified location. For example: ``Put a heated apple next to the lettuce on the middle shelf in the refrigerator."
\paragraph{Cool \& Place} This is the same as above, but with a refrigerator instead of a microwave. For example, ``Drop a cold potato slice in the sink."

\paragraph{Clean \& Place} The agent must put an object into the sink and turn on the water to clean the object. Then, it must be placed at a specified location. For example, ``Put a washed piece of lettuce on the counter by the sink."

\paragraph{Stack \& Place} The agent must pick up an object, place it into a receptacle, and then bring the stacked objects to a specified location and place them. For example, ``Move the pan on the stove with a slice of tomato in it to the table."

\paragraph{Pick Two \& Place} The agent must pickup an object, place it somewhere, then pick up another instance of that object and put it in the same place. For example, ``Place two CDs in top drawer of black cabinet."
\\\\
These last two task types, \textbf{Stack \& Place} and \textbf{Pick Two \& Place}, are the ones held out in the \emph{Novel Tasks} generalization experiments.